\documentclass[10pt,twocolumn,letterpaper]{article}

\usepackage{wacv}              % To produce the CAMERA-READY version
\usepackage{graphicx}
\usepackage{amsmath}
\usepackage{amssymb}
\usepackage{booktabs}

\usepackage{subcaption}
\usepackage{mathtools}
\usepackage{kotex}

\usepackage{algorithmicx}
\usepackage{algorithm}
\usepackage{algpseudocode}
\usepackage[normalem]{ulem}
\usepackage{ifthen}

\usepackage{xcolor}

\usepackage[pagebackref,breaklinks,colorlinks]{hyperref}

\usepackage[capitalize]{cleveref}
\crefname{section}{Sec.}{Secs.}
\Crefname{section}{Section}{Sections}
\Crefname{table}{Table}{Tables}
\crefname{table}{Tab.}{Tabs.}

\def\wacvPaperID{1302} % *** Enter the WACV Paper ID here
\def\confName{WACV}
\def\confYear{2025}

\begin{document}

%%%%%%%%% TITLE - PLEASE UPDATE
\title{Skip-and-Play: Depth-Driven Pose-Preserved Image Generation for Any Objects}

\author{Kyungmin Jo\\
KAIST\\
Daejeon, Korea\\
{\tt\small bttkm@kaist.ac.kr}% For a paper whose authors are all at the same institution,
% omit the following lines up until the closing ``}''.
% Additional authors and addresses can be added with ``\and'',
% just like the second author.
% To save space, use either the email address or home page, not both
\and
Jaegul Choo\\
KAIST\\
Daejeon, Korea\\
{\tt\small jchoo@kaist.ac.kr}
}

\twocolumn[{
\maketitle
\begin{center}
    \captionsetup{hypcap=false}
    \centering 
    \includegraphics[width=0.95\linewidth]{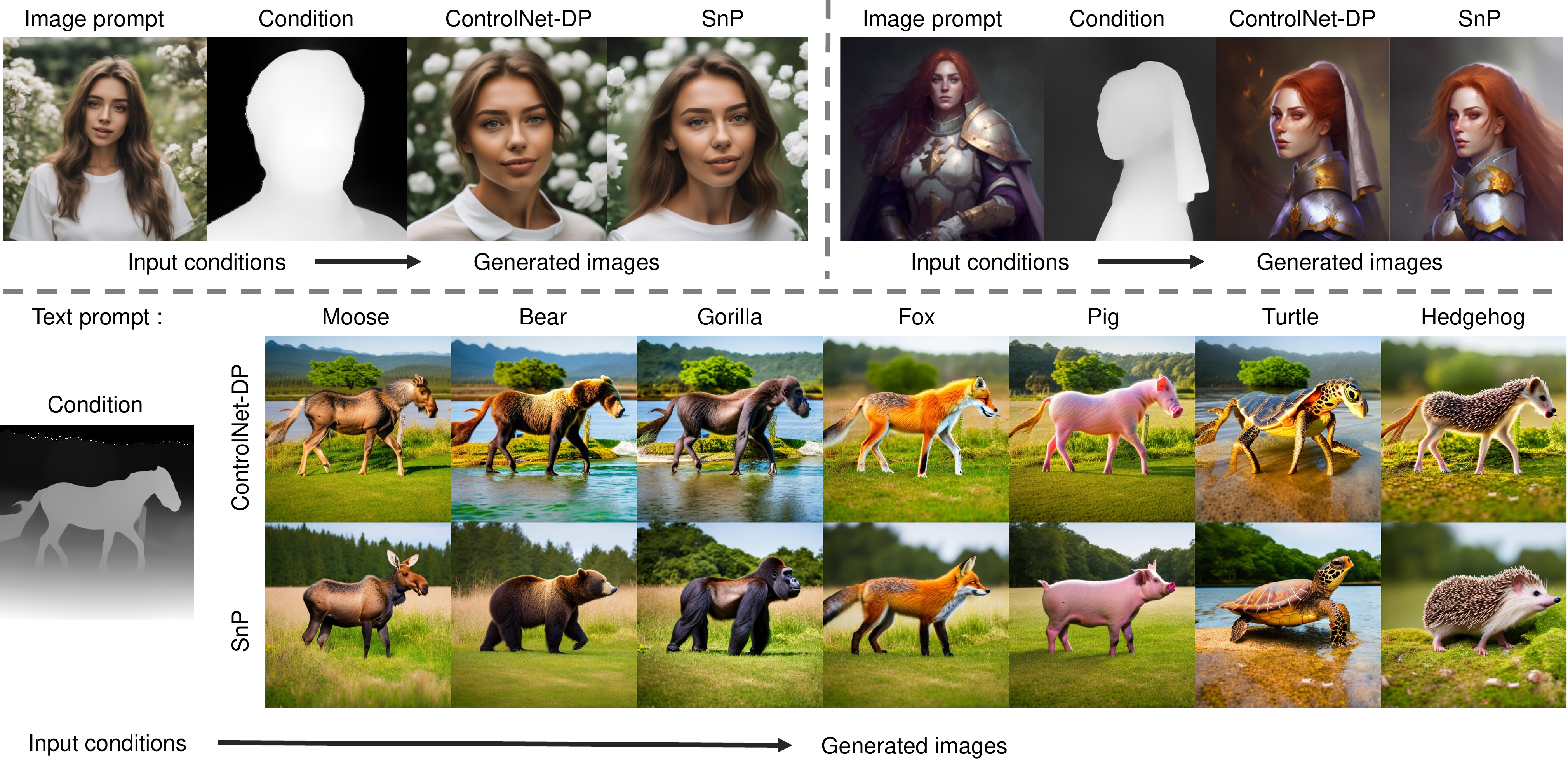}
    \vspace{-0.3cm}
    \captionof{figure}{\small Our method, Skip-and-Play (SnP), generates images of \textit{any objects} from either image prompts (top) or text prompts (bottom), \textit{reflecting the given poses} of conditions. While a depth (DP)-conditional ControlNet generates images reflecting object shapes from the condition, SnP produces images where the shapes reflect the prompt rather than the condition, despite employing the same model \textit{without additional training}. For instance, when using the prompt "pig" and the depth map of a horse image as the condition, ControlNet produces a pig with the shape of a horse, while SnP does not. Extra results and the full text prompts are in the Supplementary (Suppl.).}
    % \vspace{0.5cm}
\label{fig:teaser}
% \vspace{-0.4cm}
\end{center}
% \end{figure*} 
}]

%%%%%%%%% ABSTRACT
\begin{abstract}
   The emergence of diffusion models has enabled the generation of diverse high-quality images solely from text, prompting subsequent efforts to enhance the controllability of these models. Despite the improvement in controllability, pose control remains limited to specific objects (\eg, humans) or poses (\eg, frontal view) due to the fact that pose is generally controlled via camera parameters (\eg, rotation angle) or keypoints (\eg, eyes, nose).
   Specifically, camera parameters-conditional pose control models generate unrealistic images depending on the object, owing to the small size of 3D datasets for training. Also, keypoint-based approaches encounter challenges in acquiring reliable keypoints for various objects (\eg, church) or poses (\eg, back view). 
   To address these limitations, we propose depth-based pose control, as depth maps are easily obtainable from a single depth estimation model regardless of objects and poses, unlike camera parameters and keypoints. However, depth-based pose control confronts issues of shape dependency, as depth maps influence not only the pose but also the shape of the generated images.
   To tackle this issue, we propose Skip-and-Play (SnP), designed via analysis of the impact of three components of depth-conditional ControlNet on the pose and the shape of the generated images. To be specific, based on the analysis, we selectively skip parts of the components to mitigate shape dependency on the depth map while preserving the pose. Through various experiments, we demonstrate the superiority of SnP over baselines and showcase the ability of SnP to generate images of diverse objects and poses. Remarkably, SnP exhibits the ability to generate images even when the objects in the condition (\eg, a horse) and the prompt (\eg, a hedgehog) differ from each other.
\end{abstract}

%%%%%%%%% BODY TEXT
\section{Introduction}
\label{sec:intro}

With the advent of large-scale text-to-image diffusion models~\cite{ramesh2022hierarchical,saharia2022photorealistic,rombach2022high}, one can generate diverse high-quality images from given text. However, since these models primarily rely on text for adjusting the generated images, subsequent research has shifted focus towards enhancing their controllability by incorporating image prompts for content control~\cite{ye2023ip,wang2024instantid}, as well as extra conditions for structure or pose control~\cite{liu2023zero,hertz2022prompt,tumanyan2023plug,zhang2023adding}. 

Despite remarkable advances in the controllability of diffusion models, pose controllability remains limited, notably enabling it only on specific objects (\eg, a human) or poses (\eg, near the frontal view) due to the fact that pose is commonly controlled through camera parameters (\eg, rotation angle) or keypoints (\eg, eyes, nose).
Specifically, approaches~\cite{liu2023zero} using camera parameters for pose control generate realistic images of only a limited scope of objects compared to models~\cite{ramesh2022hierarchical,saharia2022photorealistic,rombach2022high} trained on large-scale 2D datasets~\cite{schuhmann2022laion}, primarily due to the limited objects in 3D datasets~\cite{downs2022google}. Additionally, keypoint-based pose control studies~\cite{zhang2023adding,mou2023t2i,wang2024instantid} face difficulties in applying to diverse objects and poses, stemming from the absence of reliable keypoints. For example, the difficulty of defining keypoints of the pose of churches hinders generating the image of them from keypoints. Similarly, depicting side views of humans using keypoints is complicated, often failing in the generation of side views compared to the frontal views (the fifth row in \cref{fig:qual_comp_base}). 

To enable generating images of \textit{any objects reflecting the given poses accurately}, we propose depth-based pose control for two reasons: 1) accessibility, and 2) accuracy. While obtaining camera parameters and keypoints necessitate training distinct estimation models for each class of object (\eg, human, chair), depth can be universally acquired using a single depth estimation model~\cite{ranftl2020towards} for any objects. 
Also, while keypoints lack 3D information due to their projection onto a 2D plane, depth inherently encodes 3D spatial information, making it more suitable for controlling pose (\cref{sec:depth_pose_reflection}), defined by rotations and translations in 3D space. For the same reason, depth maps are superior for pose control to other structural control conditions such as segmentation maps, edge maps, \etc.

However, since depth maps contain information not only about the pose but also about the shape, images generated using them as conditions inherit both poses and shapes of them. 
For instance, generating an image of a hedgehog guided by a depth map of a horse image results in a hedgehog with a horse-like shape (the last example of ControlNet-DP in \cref{fig:teaser}). For this reason, previous studies~\cite{zhang2023adding} have utilized depth not for pose control but for structure control.
To overcome this issue, we introduce Skip-and-Play (SnP), designed through a comprehensive analysis of the effects of three key components of ControlNet on the pose of the generated images: 1) the time steps using ControlNet, 2) the features generated from ControlNet using negative prompts, and 3) the ControlNet features passed to each decoder block. By selectively skipping a part of three elements, SnP enables the image generation of various objects reflecting the specified pose dictated by depth, without having a depth-dependent shape.

To sum up, our key contributions are as follows:
\begin{itemize}
    \item We propose utilizing depth for pose control in a diffusion model, as depth is obtainable for any objects and poses and inherently encodes 3D information, making it suitable for representing poses defined in this space.
    \item We propose Skip-and-Play, designed by the empirical insights of depth-conditional ControlNet, to generate images reflecting the given pose without the shape being dependent on the depth map. 
    \item We experimentally demonstrate the superiority of our model, both qualitatively and quantitatively, compared to previous studies on pose control in diffusion models.
\end{itemize}

\section{Related Work}
\label{sec:related_Work}

\noindent{\bf Pose-guided Image Generation.}
After the inception of Generative Adversarial Networks (GANs), a concerted effort has been made to generate images reflecting given poses.
3D GANs~\cite{niemeyer2021giraffe,chan2021pi,chan2022efficient} and 3D diffusion models~\cite{liu2023zero} directly manipulate poses by training Neural Radiance Fields~\cite{mildenhall2021nerf}-based networks using datasets composed of images and the corresponding camera parameters. %enabling the generation of images that directly reflect the desired pose. 
Unlike 3D models, there are also studies that control poses in 2D space. SeFa~\cite{shen2021closed} controls pose in pre-trained GANs by decomposing their weights. %, without additional networks or training. 
Several studies~\cite{pan2023drag,shi2023dragdiffusion} control poses of the images by moving the features of keypoints towards target positions through test-time optimization. Other approaches~\cite{bai2022single,han2018viton,zhang2023adding,yang2024imagebrush,okuyama2024diffbody,zhao2024uni} generate human images guided by estimated keypoints of the reference images obtained via keypoint detection models~\cite{cao2017realtime}. 
However, these direct pose control methods face challenges in generating realistic images or accurately reflecting poses. Specifically, training them requires datasets that pair images with corresponding camera parameters or keypoints, complicating the construction of datasets with diverse objects and resulting in unrealistic images depending on the target objects. Moreover, models that use a limited number of keypoints for pose control often struggle to achieve precise pose accuracy.

\noindent{\bf Structure-guided Image Generation.}
Unlike the pose-guided generation methods, studies have indirectly guided poses of generated images by using structures containing pose information. Diffusion-based image-to-image translation~\cite{tumanyan2023plug} and editing~\cite{hertz2022prompt} models generate new domain or style images while preserving the structure of the reference image by injecting attention from the reference into the new image. SDEdit~\cite{meng2022sdedit} adds noise to the reference image and generates an image from it through a denoising process. Also, several approaches~\cite{zhang2023adding,mou2023t2i,wang2024instantid,zhao2024uni} add networks to reflect the structure of given conditions, such as segmentation maps, edge maps, and depth maps, to the generated images. These structure-guided image generation methods can generate images of desired poses, however, they face the issue of controlling not only the pose but also the shape due to the shape information in the structural control conditions.

\noindent{\bf Image Generation from Rough Conditions.}
Recent models~\cite{bhat2023loosecontrol,liu2024smartcontrol} have emerged that generate images from rough conditions, reducing the need for precisely aligned conditions in controllable generation methods~\cite{zhang2023adding}.
LooseControl~\cite{bhat2023loosecontrol} generates images reflecting the prompt from depth maps composed of 3D boxes, rather than precise shapes of objects.
SmartControl (SC)~\cite{liu2024smartcontrol}, closely related to SnP, uses an additionally trained control scale predictor (SCP) to adjust local control scales for ControlNet feature maps. Specifically, it reduces the weights of areas conflicting between the condition and the prompt, ensuring faithful reflection of the given condition while guiding conflicting areas to reflect the prompt. 
These models are designed to generate images from rough conditions, not to control pose, thus they do not accurately reflect the pose of the condition.
To the best of our knowledge, we are the first to utilize depth for pose control in diffusion models. Despite using depth for control, we generate images with shapes reflecting the content of the prompt across various objects, surpassing previous studies (\cref{sec:exp_results}).

\section{Preliminary}
\label{sec:preliminary}

\noindent{\bf ControlNet.}
\label{sec:controlnet}
To enhance the controllability of existing pre-trained diffusion models, ControlNet~\cite{zhang2023adding} adds a ControlNet encoder $E_C$ that takes conditions $\textbf{c}_\text{i}$ (\eg, edge map) as inputs to diffusion models, which consist of the encoder $E$ and the decoder $D$ of UNet~\cite{ronneberger2015u}.
The architecture of the ControlNet encoder $E_C$ is the same as the encoder $E$, except for additional zero convolutions to the output of each block and four convolution layers for the condition $\textbf{c}_\text{i}$. 
For reflecting the condition $\textbf{c}_\text{i}$ in the generated images, ControlNet utilizes it along with the input $\textbf{z}_{t}$ at the time step $t$ and a prompt $c$ to obtain outputs $\epsilon_\theta$ as follows:
\begin{equation}
    \epsilon_\theta(\textbf{z}_{t}, t, c, \textbf{c}_\text{i}) = D(E(\textbf{z}_{t}, t, c), E_C(\textbf{z}_{t}, t, c, \textbf{c}_\text{i}))).
\label{eq:cn_epsilon}
\end{equation}
In this process, the features generated from the ControlNet encoder $E_C$ are added to the corresponding features from the encoder $E$ before passing to the decoder $D$.
In the case of applying classifier-free guidance~\cite{nichol2021glide}, two outputs $\epsilon_\theta^{+}$ and $\epsilon_\theta^{-}$ are estimated using the positive $c^+$ and negative prompts $c^-$, respectively, as follows:
\begin{equation}
    \epsilon^+_\theta(\textbf{z}_{t}, t, c^+, \textbf{c}_\text{i}) = D(E(\textbf{z}_{t}, t, c^+), E_C(\textbf{z}_{t}, t, c^+, \textbf{c}_\text{i}))),
\label{eq:cn_pos_epsilon}
\end{equation}
\begin{equation}
    \epsilon^-_\theta(\textbf{z}_{t}, t, c^-, \textbf{c}_\text{i}) = D(E(\textbf{z}_{t}, t, c^-), E_C(\textbf{z}_{t}, t, c^-, \textbf{c}_\text{i}))),
\label{eq:cn_neg_epsilon}
\end{equation}
where the positive $c^+$ and negative prompts $c^-$ refer to the conditions to be included and excluded, respectively, in the generated image.
Using two outputs, the final output $\epsilon_\theta$ is defined as: \begin{equation}
\begin{aligned}
\epsilon_\theta(\textbf{z}_{t}, t, c^{+}, c^{-}, \textbf{c}_\text{i}) &= \epsilon_\theta^{-}(\textbf{z}_{t}, t, c^{-}, \textbf{c}_\text{i}) \\
&+ s \cdot (\epsilon_\theta^{+}(\textbf{z}_{t}, t, c^{+}, \textbf{c}_\text{i}) - \epsilon_\theta^{-}(\textbf{z}_{t}, t, c^{-}, \textbf{c}_\text{i})),
\label{eq:cfg_noise}    
\end{aligned}
\end{equation}
where $s$ is the guidance scale with a value greater than 1.
\begin{figure}[!t]
    \centering
    \begin{subfigure}[b]{0.55\linewidth}
        \centering
        \includegraphics[width=\linewidth]{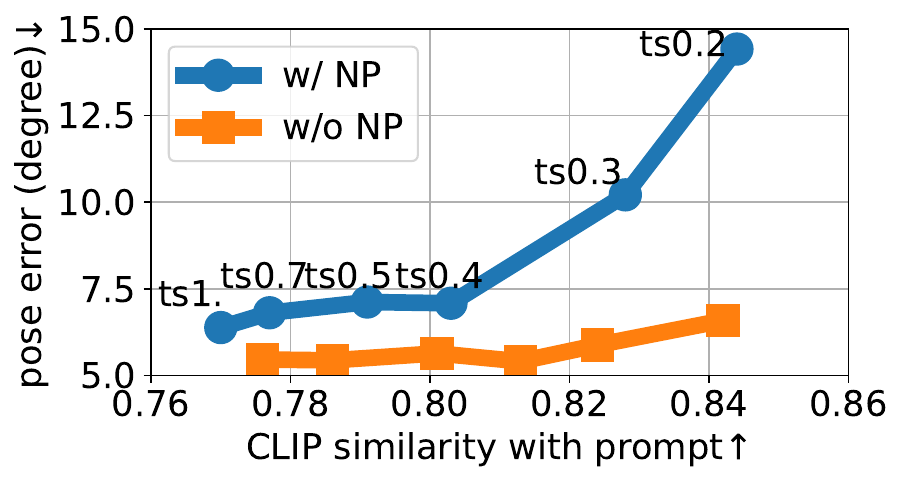}
        \caption{Negative prompt (NP)}
        \label{fig:skip_neg}
    \end{subfigure}
    \begin{subfigure}[b]{0.42\linewidth}
        \centering
        \includegraphics[width=\linewidth]{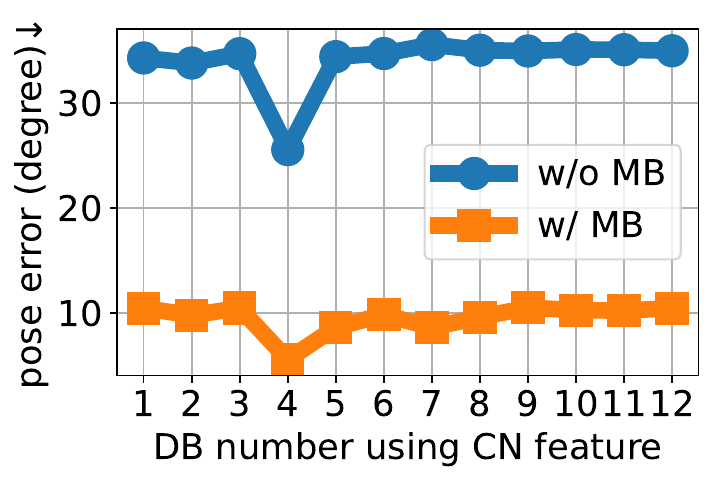}
        \caption{Each decoder block (DB)}
        \label{fig:skip_block}
    \end{subfigure}
    \vspace{-1mm}
    \caption{Impact of three components of ControlNet on the pose of the generated images. (a) Impact of NP in the ControlNet encoder. With NP, using ControlNet up to 0.4 time steps leads to a notable decrease in pose error between the generated images and conditions, but using it beyond this step yields marginal improvement in pose error. However, not using NP aids in reflecting the given pose across time steps using ControlNet. ts indicates $\lambda_{t}$. (b) Among the ControlNet features, features for the middle block (MB) and the fourth DB have the most significant impact on the pose.}
    \label{fig:skip_graph}
\vspace{-2mm}
\end{figure}

\begin{figure}[!t]
\includegraphics[width=\linewidth]{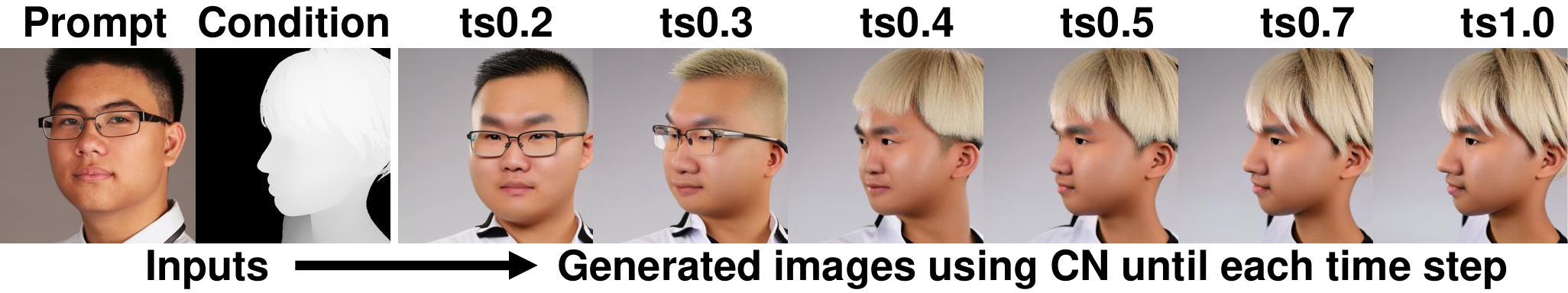}
% \vspace{-0.2cm}
\caption{Visual results according to the time steps $\lambda_{t}$ using ControlNet in the blue line in \cref{fig:skip_neg}. Using ControlNet up to 0.4 time steps reflects the pose of a given condition, but the shape of the condition is also reflected in the generated image.
}
\label{fig:qual_skip_ts}
\vspace{-5mm}
\end{figure}

\section{Method}
\label{sec:method_snp}
We elucidate the methodology for generating images that reflect the poses of the conditions and the contents of prompts. 
To reflect the pose of the conditions, we adopt depths for two reasons: 1) accessibility, and 2) accuracy. Specifically, depths are easily obtainable for any objects and poses using a single depth estimation model~\cite{ranftl2020towards}, unlike camera parameters or keypoints. Additionally, unlike 2D projected keypoints, depths inherently encode 3D spatial information, enabling more precise control of poses defined in 3D space (\cref{sec:depth_pose_reflection}).
For depth-conditional image generation, we adopt ControlNet~\cite{zhang2023adding} based on Stable Diffusion (SD)~\cite{rombach2022high} as a baseline to reflect the pose of the given condition.

In this section, we first provide an analysis of depth-conditional ControlNet in~\cref{sec:analysis_cn}, followed by an explanation of SnP designed based on this analysis (\cref{sec:skipNplay}). 
For the experiments in this section, we utilize the IP-Adapter~\cite{ye2023ip} to employ image prompts, aiming to discern whether the characteristics of the generated images originate from the prompt or the condition. Although we use image prompts for analysis, our approach is not restricted to image prompts and can also utilize text prompts (\cref{fig:teaser}). 

\subsection{Analysis of ControlNet on the Pose of Image}
\label{sec:analysis_cn}
Depths provide information not only about the pose but also about the shape, resulting in depth-dependent shapes in images generated by depth-conditional ControlNet (\cref{fig:teaser}). 
To mitigate this problem and reflect contents including the shapes from the prompts (the results of SnP in \cref{fig:teaser}), 
%Yet, our goal is to generate images that maintain the poses of the conditions while reflecting contents including the shapes from the prompts (the results of SnP in \cref{fig:teaser}).
inspired by~\cite{tumanyan2023plug}, we thoroughly analyze the influence of three components of ControlNet on the pose of the generated images: 1) time step using ControlNet, 2) ControlNet features generated using the negative prompt (NP), and 3) ControlNet features passed to each decoder block (DB).

%%%%%%%%%%%%%%%%%%%%%%%%%%%%%%%%%%%%%%%%%%%%%%%%%%%%%%
\paragraph{Time Steps using ControlNet.}
\label{sec:skip_ts}
Since the shape of the generated image is determined during the initial time steps~\cite{hertz2022prompt}, the simplest way to minimize the influence of depths on the shape of the generated images is to halt the use of ControlNet at early time steps as follows:
\begin{equation}
\epsilon_\theta(\textbf{z}_{t}, t, c, \textbf{c}_\text{i}) = \begin{cases*}
  \epsilon_\theta(\textbf{z}_{t}, t, c, \textbf{c}_\text{i}), 
  & if $t \leqq \lambda_{t}$ ,\\
  \epsilon_\theta(\textbf{z}_{t}, t, c),                    & otherwise,
\end{cases*}
\label{eq:cfg_ts}
\end{equation}
where $\lambda_{t}$ is a threshold of time steps using ControlNet. 
As depicted by the blue line in \cref{fig:skip_neg}, the pose error exhibits different patterns depending on whether the last time step (ts) using ControlNet is below or over 0.4.
Specifically, if ControlNet is used beyond this point, the pose error decreases slightly, but depths not only affect the pose of the generated image but also has a significant impact on their shapes (\cref{fig:qual_skip_ts}). Conversely, if we halt the use of ControlNet before this point, the generated image adopts a shape akin to the prompt rather than the depth map (\cref{fig:qual_skip_ts}), but the pose of the generated image deviates from that of the depth map (the blue line in \cref{fig:skip_neg}).
This indicates that both the pose and shape are simultaneously affected and altered by depths in ControlNet, thus merely adjusting the time steps for applying ControlNet does not generate images that reflect both the pose from the depth map and the shape from the prompt.
However, although adjusting the time steps using ControlNet is insufficient for reflecting the pose and shape in the generated image from the depth map and prompt, respectively, ceasing the use of ControlNet early enough can mitigate the effect of depth on the shape of the generated images (\cref{fig:qual_skip_ts}).
Thus, to decrease the impact of depth on the shape of images, in SnP, we control the usage of ControlNet based on time steps to ensure that ControlNet features are applied until early time steps.
Nevertheless, this leads to the pose of the depth map not being accurately reflected in the generated images, as previously mentioned. To address this issue, we shift our attention to the ControlNet features generated from the negative prompt.
%%%%%%%%%%%%%%%%%%%%%%%%%%%%%%%%%%%%%%%%%%%%%%%%%%%%%%
\begin{figure}[!t]
\includegraphics[width=\linewidth]{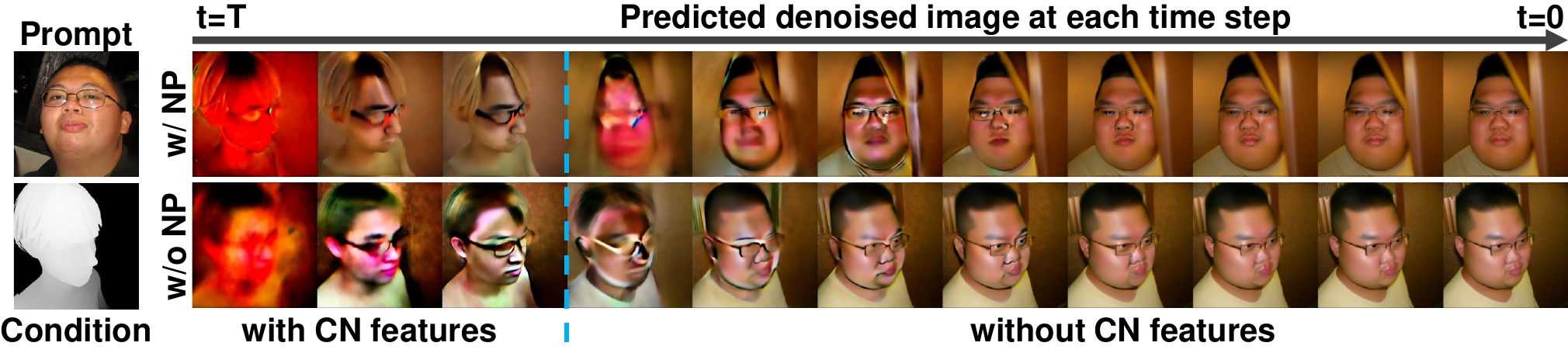}
% \vspace{-0.2cm}
\caption{Visual results of predicted denoised images at each time step with (top) and without (bottom) using ControlNet features from NP. These images depict the visual outcomes of ts0.2 in \cref{fig:skip_neg}. Utilizing ControlNet features from NP causes a change in the pose of the image at the moment of cessation of ControlNet usage (blue dashed line) and creates shapes dependent on the depth map when using ControlNet.
}
\label{fig:qual_skip_neg}
\vspace{-5mm}
\end{figure}

\paragraph{ControlNet Features Obtained from Negative Prompt.}
\label{sec:skip_neg}
According to ControlNet~\cite{zhang2023adding}, removing the feature maps $E_C^-=E_C(\textbf{z}_{t}, t, c^-, \textbf{c}_\text{i})))$ obtained from the ControlNet encoder $E_C$ using a negative prompt, boosts the reflection of conditions $\textbf{c}_\text{i}$ in the generated images. 
Taking it one step further, we have found that eliminating $E_C^-$ enhances the reflection of the poses of the condition without compromising the reflection of the prompt in the generated images regardless of the time steps $\lambda_t$ using ControlNet (the orange line in \cref{fig:skip_neg}). For example, when ControlNet is used up to 0.2 time steps, utilizing $E_C^-$ results in an average pose error of 14.42 degrees, whereas removing $E_C^-$ lowers the pose error to 6.58 degrees. On the other hand, the content reflection evaluated based on the CLIP cosine similarity is similar in both cases.
The effects of removing $E_C^-$ on the pose of the generated images can be explained by comparing the noise estimation process of classifier-free guidance in terms of the usage of $E_C^-$.
Compared to the outputs estimated \textit{using} $E_C^-$ in~\cref{eq:cfg_noise}, the outputs estimated \textit{without using} $E_C^-$ is calculated as
\begin{equation}
\begin{aligned}
\epsilon_\theta(\textbf{z}_{t}, t, c^{+}, c^{-}, \textbf{c}_\text{i}) &= \epsilon_\theta^{-}(\textbf{z}_{t}, t, c^{-}) \\
&+ s \cdot (\epsilon_\theta^{+}(\textbf{z}_{t}, t, c^{+}, \textbf{c}_\text{i}) - \epsilon_\theta^{-}(\textbf{z}_{t}, t, c^{-})).
\label{eq:cfg_woneg}
\end{aligned}
\end{equation}
According to GLIDE~\cite{nichol2021glide}, the classifier-free guidance can be interpreted as moving the output of each time step away from $\epsilon_\theta^{-}$ towards the direction of $\epsilon_\theta^{+}$. 
Based on this explanation, we can intuitively elaborate on the effect of removing $E_C^-$ on the reflection of conditions. 
When \textit{using} $E_C^-$, in \cref{eq:cfg_noise}, the condition $\textbf{c}_\text{i}$ is applied to the generated images along with the negative prompt $c^{-}$ in the first term on the right-hand side, and in the next term, the output moves in the direction from applying $c^{-}$ to $c^{+}$. 
Conversely, in \cref{eq:cfg_woneg}, \textit{removing} $E_C^-$, the output moves in the direction from applying $c^-$ to simultaneously applying both $\textbf{c}_\text{i}$ and $c^{+}$, with $s$ amplifying this movement. 
Thus, $\textbf{c}_\text{i}$ and $c^{+}$ are more jointly and rapidly applied to the generated images when removing $E_C^-$ compared to using it.
This tendency is also apparent in the visual results when $E_C^-$ is utilized and omitted. In \cref{fig:qual_skip_neg}, the images depict the denoised image predicted at each time step, with applying ControlNet until 0.2 time step. When comparing the outcomes before halting the use of ControlNet (images on the left of the blue dashed line), the removal of $E_C^-$ (bottom) benefits a smooth integration of pose and prompt reflection. In contrast, the use of $E_C^-$ (top) yields precise pose reflection but insufficient prompt reflection, leading to depth-dependent shape issues. %leading to shapes resembling those under the depth condition.
Furthermore, removing $E_C^-$ ensures pose consistency even after terminating the use of ControlNet.

\begin{figure}[!t]
\includegraphics[width=\linewidth]{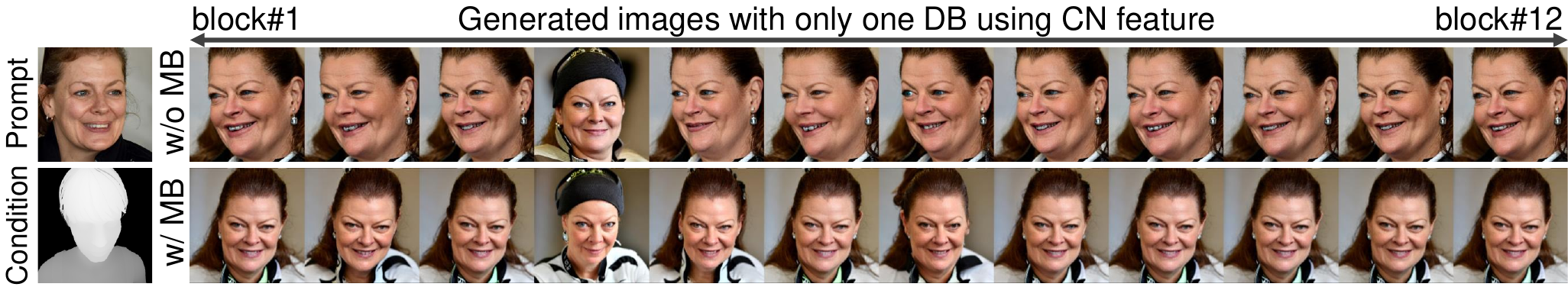}
% \vspace{-0.2cm}
\caption{
Generated images using ControlNet features in each decoder block (DB) at a time (Top: without ControlNet features in the middle block (MB), Bottom: with the features in the MB). These correspond to the blue and orange lines in \cref{fig:skip_block}, respectively. ControlNet features added to the MB control coarse pose, while those added to the features for the fourth DB adjust fine pose and image shape.}
\label{fig:qual_skip_block}
\vspace{-5mm}
\end{figure}
% %%%%%%%%%%%%%%%%%%%%%%%%%%%%%%%%%%%
\paragraph{ControlNet Features for Each Decoder Block.}
\label{sec:skip_db}
We assess the impact of each feature map generated from every block in the ControlNet encoder $E_C$ on the pose of the images and have found that only a subset of blocks significantly influence the pose of the generated images.
Specifically, we generate images using only the feature map of one block at a time and compare the pose error between the generated images and depth maps. Also, we divide the evaluation into two cases (\cref{fig:skip_block}): one where the features of the middle block (MB) are used (orange line) and the other where they are not used (blue line).
As a result, only two blocks—specifically, the MB and the block corresponding to the fourth decoder block—influence the pose of the generated images. 
To be specific, the MB has the most significant impact on the pose of the generated images, followed by the fourth block in the decoder. The remaining blocks have minimal influence on the pose. Also, as shown in \cref{fig:qual_skip_block}, the MB only impacts the pose, whereas the fourth block impacts both the pose and the shape.
According to our analysis, the blocks that influence the pose of the generated images vary depending on the baseline model and are independent of the type of condition. Refer to the Suppl. for more details.

%%%%%%%%%%%%%%%%%%%%%%%%%%%%%%%%%%%%%%%%%%%%%%%%%%%%%%%%%%
\begin{figure}[!t]
\includegraphics[width=\linewidth]{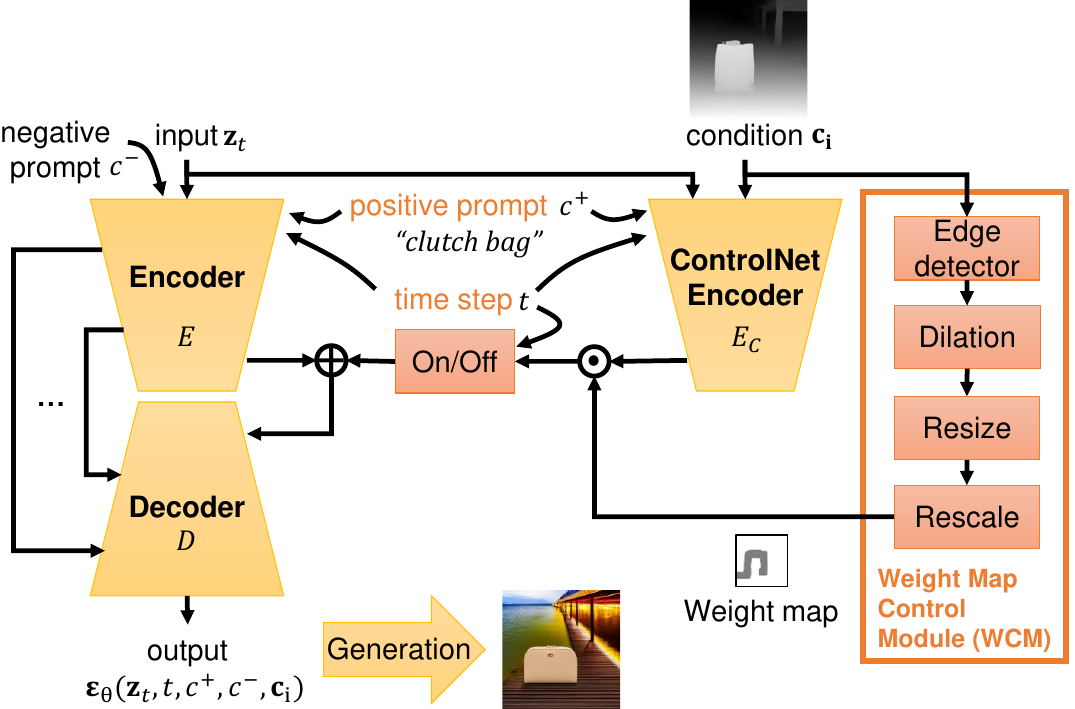}
% \vspace{-0.2cm}
\caption{The architecture of Skip-and-Play.
}
\label{fig:architecture}
\vspace{-7mm}
\end{figure}

\subsection{Skip-and-Play}
\label{sec:skipNplay}
Based on the empirical insights obtained via the analysis (\cref{sec:analysis_cn}), we propose a new approach called Skip-and-Play (SnP) for pose-preserved image generation for any objects by reducing the influence of the depth on the shapes of generated images.
As shown in~\cref{fig:architecture}, we skip on a part of the three components in ControlNet explained in~\cref{sec:analysis_cn}. 
Specifically, to minimize influence of the depth condition on aspects other than the pose of the generated images, we apply ControlNet features to the pose-related DB and use ControlNet up to $\lambda_t$. Also, we use NP only for the encoder $E$ to accurately reflect the pose of depth maps in the generated images even in the early time steps. 

In addition, we optionally apply the Weight Map Control Module (WCM) to reduce the influence of the depth maps on the shape of objects in the generated images. The WCM detects edges of the depth map and assigns lower weights to these areas to minimize their impact on shape. Specifically, we use an edge detector~\cite{canny1986computational} on the depth condition to identify edges, then expand these edges through dilation and invert them. Since depth maps, unlike images, are smoothed and lack fine details, this process effectively identifies the boundaries between objects and the background. Next, we resize the results to match the resolution of ControlNet features and rescale the values to ensure they fall within a specific range. Our analysis indicates that applying weights above a certain threshold to ControlNet features minimizes their impact on pose while primarily influencing shape. Thus, we adjust the weight maps accordingly before applying them to the ControlNet features. Refer to the Suppl. for more details.

\section{Experimetal Results}
\label{sec:exp_results}

In this section, we delve into our experimental findings. We begin by substantiating the superiority of SnP through both quantitative and qualitative comparisons with pose-guided and rough conditional image generation models in \cref{sec:comp_baselines}. 
Following that, we compare the performance of SnP with methods that indirectly control pose via structure (\cref{sec:comp_pnp}). 
Also, we show that despite utilizing a depth as a conditioning factor, SnP generates images with shapes more closely aligned with the prompts than depth conditions (\cref{sec:various_obj}).
In \cref{sec:ablation}, we conduct ablation studies based on combinations of components in SnP and show validity of SnP not only on SD~\cite{rombach2022high} used in our analysis but also on SDXL~\cite{podell2023sdxl}.
Lastly, in \cref{sec:depth_pose_reflection}, we show the superiority of depth-based pose control over keypoint-based pose control.
Refer to the Supple. for additional qualitative results, experimental settings, implementation details.

\begin{figure}[!t]
\centering
\includegraphics[width=\linewidth]{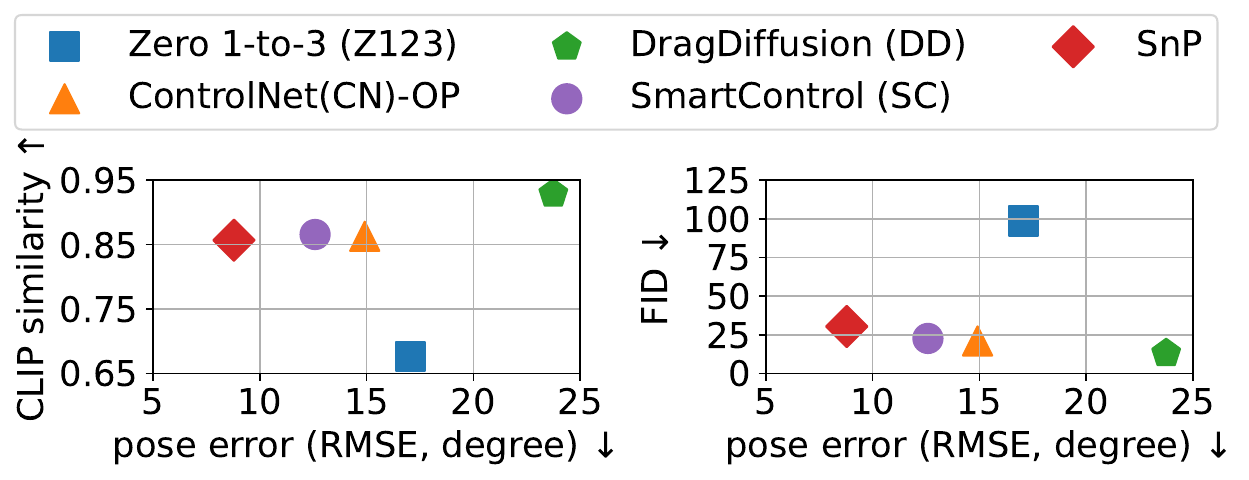}
% \vspace{-0.2cm}
\caption{Quantitative Comparison of direct pose control between SnP and baselines. 
We display (left) pose error and CLIP similarity to evaluate the given pose and prompt reflection (closer to the top left indicates better performance), and (b) pose error and FID to evaluate the given pose reflection and photorealism of images (closer to the bottom left indicates better performance). SnP surpasses methods by generating images that best reflect the pose while also producing realistic images reflecting the prompt.}
\label{fig:quan_baslines}
\vspace{-5mm}
\end{figure}

\subsection{Comparison of Direct Pose Control}
\label{sec:comp_baselines}
To show the superiority of SnP, we compare the quantitative and qualitative results of SnP to those of four baseline models: Zero 1-to-3 (Z123)~\cite{liu2023zero}, DragDiffusion (DD)~\cite{shi2023dragdiffusion}, OpenPose (OP)~\cite{cao2017realtime} conditional ControlNet (CN)~\cite{zhang2023adding}, and SmartControl (SC)~\cite{liu2024smartcontrol}. 
Our goal is to generate images reflecting the given pose, we select three diffusion models that directly control pose for image generation as baselines. Zero-1-to-3 controls pose using camera parameters, while DragDiffusion and ControlNet control pose using keypoints. 
Additionally, we utilize SC, which generates images from rough conditions, as a baseline. Although it does not aim to directly control pose, it reflects conditions by reducing ControlNet feature weights only in areas that conflict with the prompt. This aligns with the concept of generating images that reflect the pose of the given conditions and the content of the prompt, making it suitable as a baseline. For a fair comparison, we use depth as the input condition for SC.

Since Zero-1-to-3 and DragDiffusion focus on altering the pose of a given image, for a fair comparison, we employ image prompts for ControlNet, SmartControl, and SnP. Furthermore, since OpenPose-conditioned ControlNet only targets humans, we evaluate models utilizing the human face dataset, FFHQ~\cite{karras2019style}. However, since in-the-wild datasets often consist of images that are mostly biased toward frontal poses and have narrow pose ranges, we construct the PoseH dataset from images rendered with a uniform pose distribution from a single 3D mesh to evaluate pose reflection across various angles. Refer to the Suppl. for details about datasets.

\begin{figure}[!t]
\includegraphics[width=\linewidth]{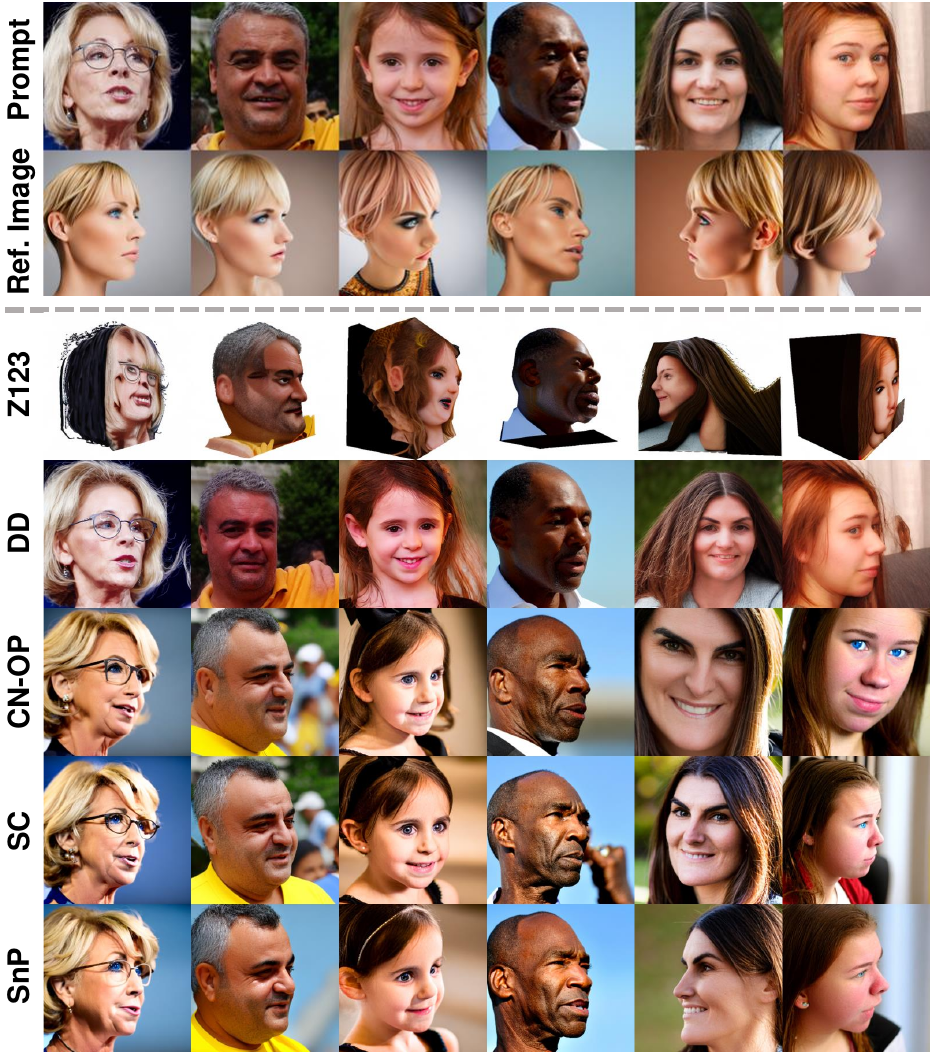}
% \vspace{-0.2cm}
\caption{Qualitative comparison of direct pose control to baselines. While baselines (from the third to the sixth rows) struggle to generate realistic images of the given pose (the second row), SnP does not encounter such difficulty. Images in the first row indicate image prompts for ControlNet (CN), SmartControl (SC), and SnP, and input images for Zero-1-to-3 (Z123) and DragDiffusion (DD). We use the image prompts for evaluation since Z123 and DD are targets to change the pose of given images.
For a fair comparison, we use the same latent for CN-OP, SC, and SnP.
}
\label{fig:qual_comp_base}
\vspace{-5mm}
\end{figure}

\subsubsection{Quantitative Comparison.}
\label{sec:quant_baseline}
The quantitative comparison is based on three metrics: a pose error, CLIP cosine similarity~\cite{radford2021learning}, and Frechet Inception Distance (FID)~\cite{heusel2017gans}. We calculate the pose error between the ground truth pose and the estimated pose of generated images from the off-the-shelf pose estimation model~\cite{hempel20226d}.
As depicted in \cref{fig:quan_baslines}, despite controlling pose using depths, SnP excels at accurately reflecting the given poses of conditions compared to all baselines, especially models directly controlling pose.
This highlights the advantage of leveraging depths for controlling poses defined in 3D space, in contrast to 2D keypoint-based pose control methods such as DragDiffusion~\cite{shi2023dragdiffusion} and ControlNet-OP~\cite{zhang2023adding}, which aligns with the results in \cref{sec:depth_pose_reflection}. Zero-1-to-3~\cite{liu2023zero} directly controls pose via camera parameters, which leads to high pose accuracy expectations. However, due to training on a limited 3D dataset, it fails to generate realistic images, resulting in degraded pose estimation performance.
SmartControl exhibits lower pose errors than other baselines by adopting depth for condition. However, its training on a small dataset occasionally leads to failures in preserve pose accurately, leading to higher pose errors compared to the training-free SnP.

\subsubsection{Qualitative Comparison.}
\label{sec:qual_baseline}
We also compare SnP to baselines qualitatively in \cref{fig:qual_comp_base}, which aligns with the results in \cref{fig:quan_baslines}. Specifically, Zero-1-to-3 generates the most unrealistic images due to training on a 3D dataset containing limited objects. On the other hand, DragDiffusion uses LoRA~\cite{hu2022lora}, allowing it to create the most realistic images reflecting the image prompts, but pose control via moving points is ineffective, especially when the distance between the poses of the given image and the target is far. ControlNet-OP can generate photorealistic images of a given pose, but, in cases like side views, it creates images with completely different poses due to the failure of OP detection (the fifth and sixth column in \cref{fig:qual_comp_base}). Like ControlNet-OP, SmartControl fails to maintain the pose of the condition in some cases as it reflects the pose of the image prompt. 
In contrast to baselines, our proposed model generates pose-preserved photorealistic images reflecting the image prompt. 

%%%%%%%%%%%%%%%%%%%%%%%%%%%%%%%%%%%%%%%%%%%%%%%%%%%%%%%%%%%%
\begin{figure}[!t]
\includegraphics[width=\linewidth]{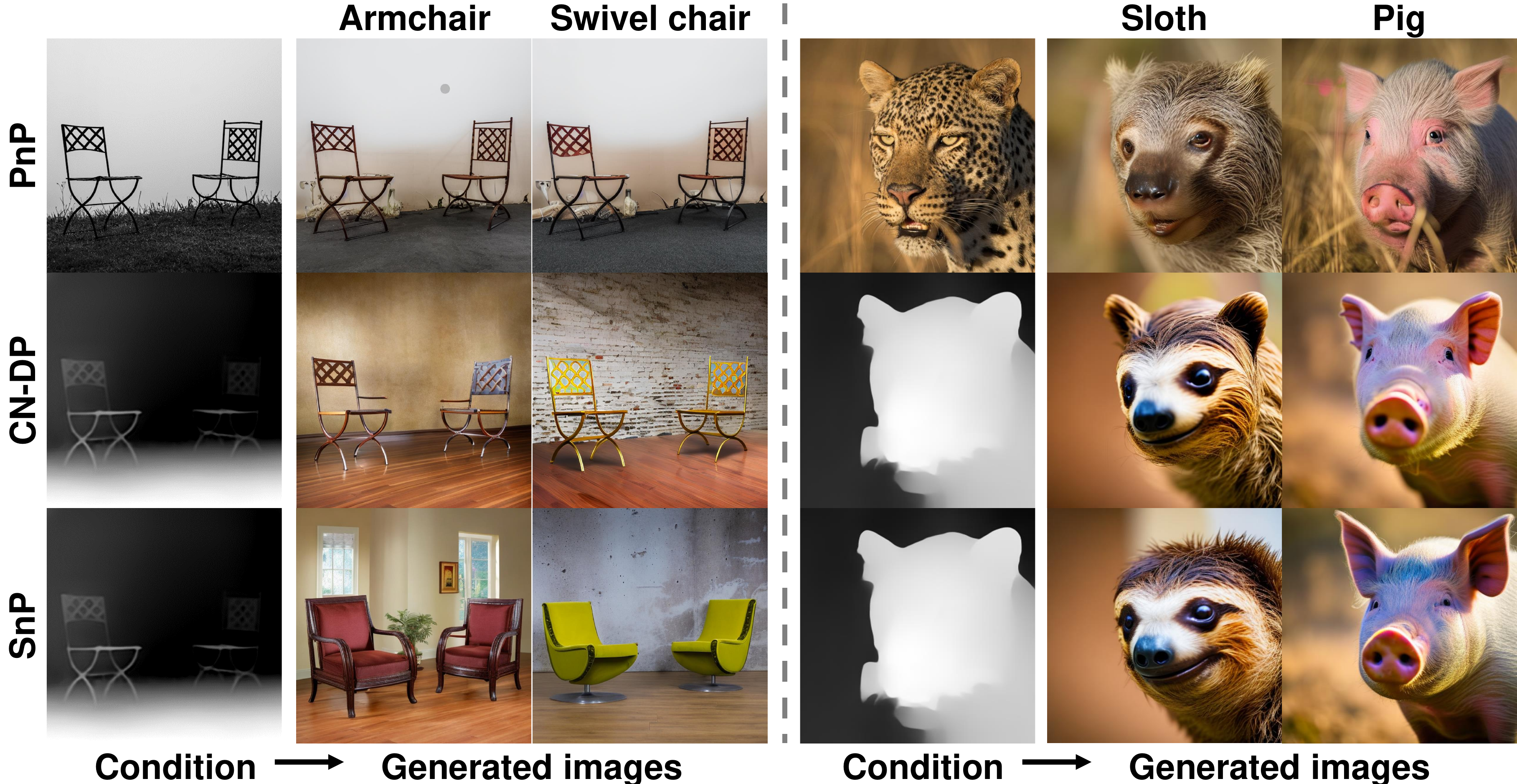}
% \vspace{-0.2cm}
\caption{Qualitative comparison to the structure-guided pose control models (PnP, ControlNet-DP). The images generated by structure-based models have shapes more closely resembling the conditions (\eg, ears) compared to the results of SnP, as they control the structure while SnP controls the pose through the conditions.
}
\label{fig:qual_comp_struct_baseline}
\vspace{-5mm}
\end{figure}

\subsection{Comparison to Structure-based Pose Control}
\label{sec:comp_pnp}
In this section, we compare the performance of SnP with structure-guided image generation models, namely Plug-and-Play (PnP) and ControlNet (CN) conditioned depth (DP). Unlike the aforementioned studies, these models that generate images by controlling structure do not aim at controlling poses, and there are no restrictions on target objects. Therefore, rather than comparing the pose accuracy for specific objects, we qualitatively compare SnP with these models across various objects. As depicted in \cref{fig:qual_comp_struct_baseline}, structure-guided image generation models, as mentioned earlier, reflect both pose and shape from the condition to the generated images. Hence, the generated images resemble the shape of the given condition more than the given prompt. 
For example, PnP and ControlNet-DP struggle to generate various chair images because they rely on the structure within the given condition.
Furthermore, images generated by both PnP and ControlNet-DP using the face of a leopard as the reference consistently feature ears resembling those of the leopard, irrespective of the species of the target animal.
On the other hand, SnP controls poses using depth conditions but reduces the dependence of shapes on these conditions, resulting in images that reflect the given prompts in shape while maintaining the poses from the depth conditions.

%%%%%%%%%%%%%%%%%%%%%%%%%%%%%%%%%%%%%%%
\begin{figure}[!t]
\includegraphics[width=\linewidth]{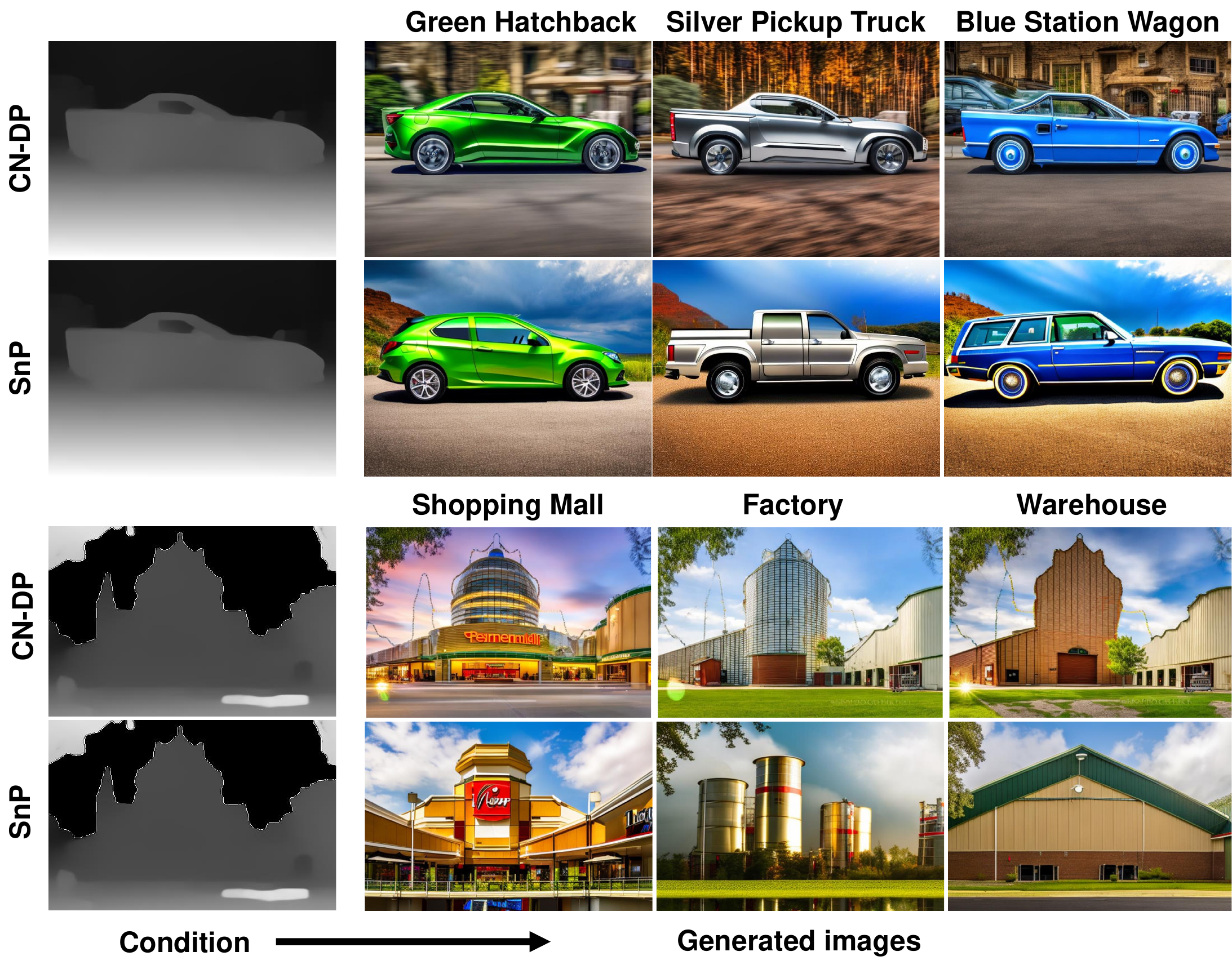}
% \vspace{-0.2cm}
\caption{Effects of SnP on the shape of generated images. We utilize the same latents and the prompts for two models for a fair comparison.
The results of the ControlNet-DP have overly depth-dependent shapes while SnP generates images with shapes according to the prompt while reflecting the given pose of the depth condition.
}
\label{fig:qual_various_obj}
\vspace{-5mm}
\end{figure}

\subsection{Effects on the Shape of Generated Images}
\label{sec:various_obj}
Compared to ControlNet-DP, SnP generates images having shapes affected more by the prompt than by the depth condition. To reveal the effectiveness of SnP, we compare the qualitative results of it and ControlNet-DP on various objects. Specifically, we sample the reference images from two datasets~\cite{krause20133d,yu2015lsun} consisting of car and church images, respectively, and generate images using depth conditions extracted from these reference images and various text prompts. As described in \cref{fig:qual_various_obj}, while ControlNet-DP generates images with shapes similar to the condition, images generated by SnP reflect the pose from the condition but have the shape more influenced by the prompt than by the condition.

%%%%%%%%%%%%%%%%%%%%%%%%%%%%%%%%%%%%%%%

\subsection{Ablation studies}
\label{sec:ablation}
\begin{figure}[!t]
    \centering
    \begin{subfigure}[b]{0.48\linewidth}
        \centering
        \includegraphics[width=\linewidth]{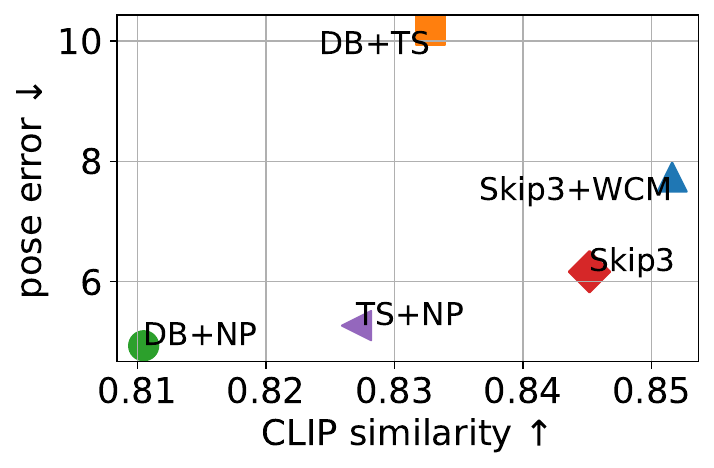}
        \caption{SD 1.5}
        \label{fig:ab_sd15}
    \end{subfigure}
    \begin{subfigure}[b]{0.48\linewidth}
        \centering
        \includegraphics[width=\linewidth]{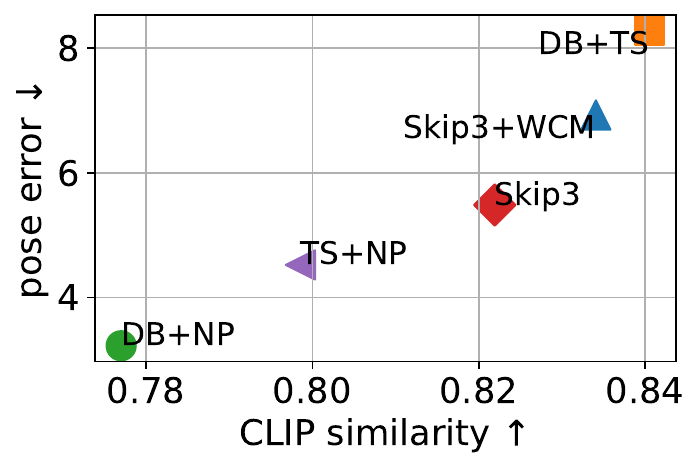}
        \caption{SDXL 1.0}
        \label{fig:ab_sdxl}
    \end{subfigure}
    \caption{Ablation studies of four components (\cref{sec:method_snp}) and backbone models. Closer to the bottom right indicates better performance. SD and SDXL exhibit similar trends in results, except when combining DB and TS. However, both models achieve the best performance when DB, TS, and NP are applied (Skip3).}
    \label{fig:quant_ablation}
\vspace{-3mm}
\end{figure}

\begin{figure}[!t]
    \centering
    \includegraphics[width=\linewidth]{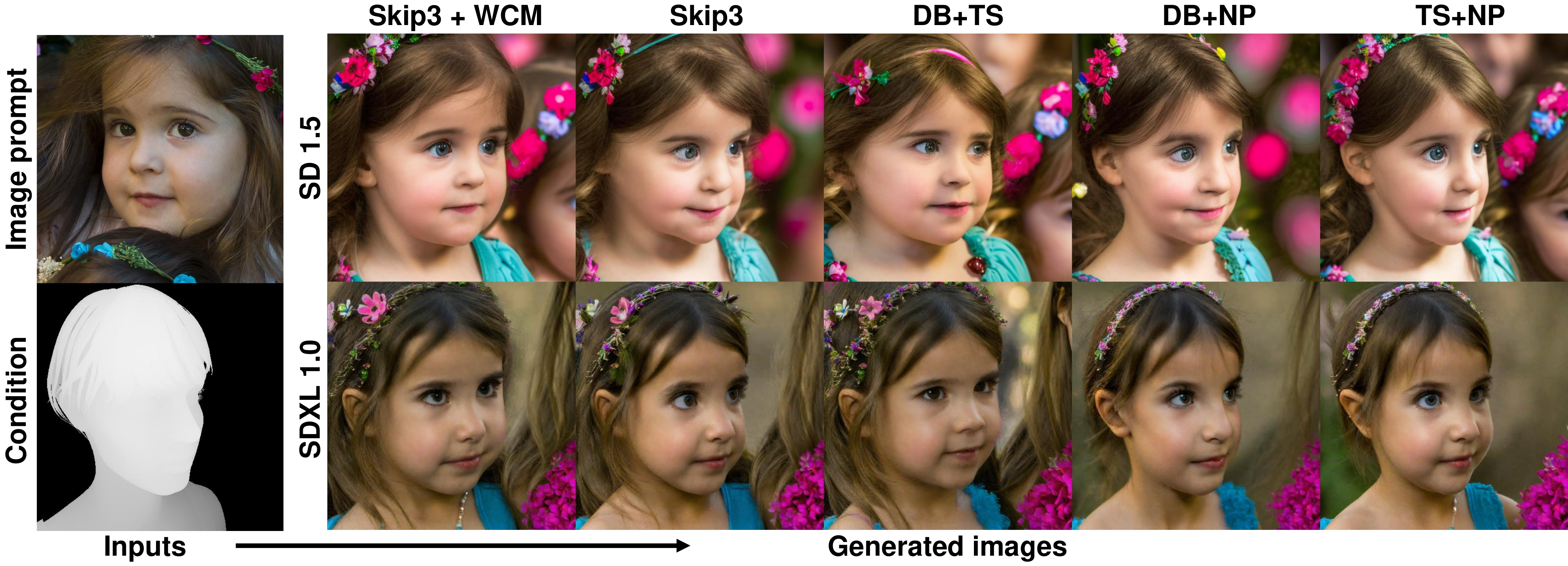}
    \caption{Visual results of ablation studies in~\cref{fig:quant_ablation}. With both models, regardless of the combination, NP benefits pose reflection, while DB and TS aid in prompt reflection. Also, WCM slightly compromises pose but enhances prompt reflection.}
    \label{fig:qual_ablation}
\vspace{-5mm}
\end{figure}

We conduct ablation studies on the baseline models and the combination of four components of SnP: 1) time steps (TS) using CN, 2) CN features generated from negative prompts (NP), 3) CN features passed to each decoder block (DB), and 4) Weight Map Control Module (WCM).
We evaluate models based on the pose error and CLIP scores to assess pose and prompt reflection, respectively. 
In the results of SD in~\cref{fig:ab_sd15}, even combined with other components, NP and TS still positively influence pose and prompt reflection, respectively. Comparing the results of using all three components (Skip3) with TS+NP, DB slightly compromises pose but positively affects prompt reflection. Additionally, the optionally applied WCM shows a similar trend as DB. These results are also evident in the visual outcomes (\cref{fig:qual_ablation}). Furthermore, we conduct the same experiment with SDXL, and the results, excluding those of DB+TS, show a similar trend to SD 1.5. With both models, applying three components yields the best performance.

%%%%%%%%%%%%%%%%%%%%%%%%%%%%%%%%%%%%%%%%%%%%%%%%%%%%%%%%
\subsection{Effects of Depth on Pose Reflection}
\label{sec:depth_pose_reflection}

\begin{figure}[!t]
  \centering
  \begin{tabular}{ c }
    \includegraphics[width=.35\linewidth]{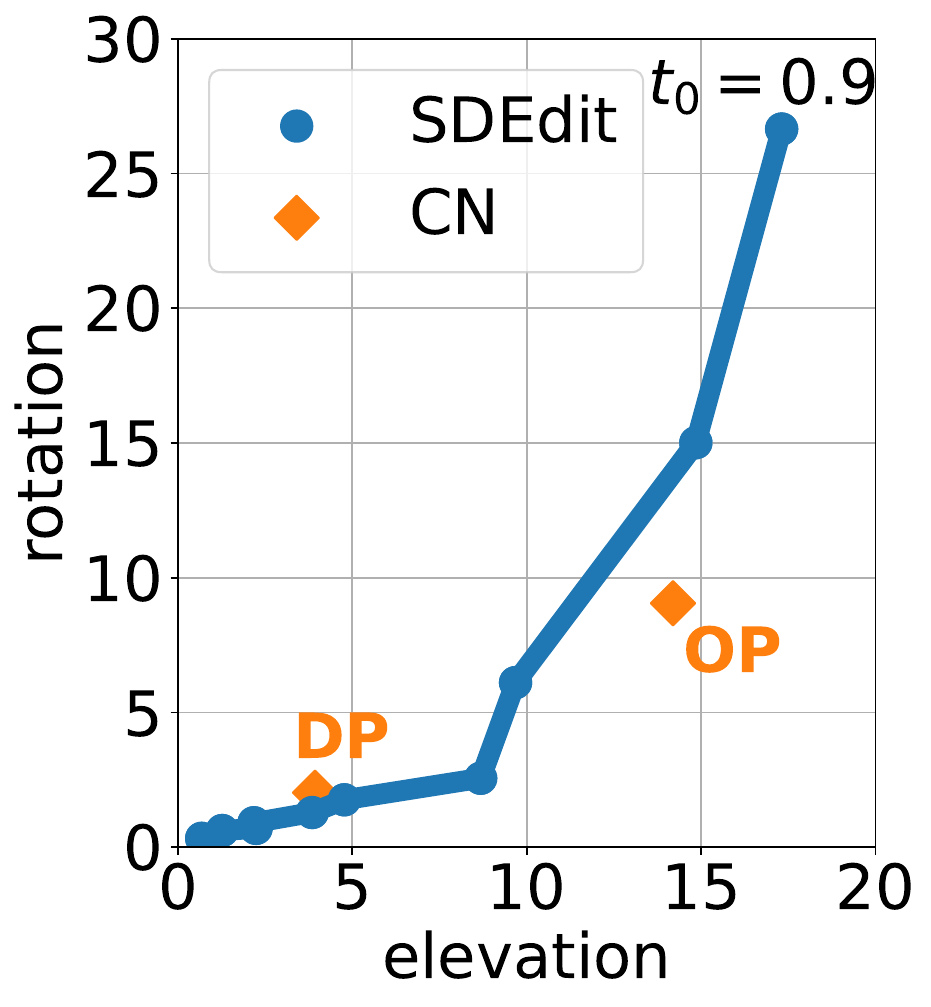}
  \end{tabular}%
  \begin{tabular}{ c c }
    \includegraphics[width=.6\linewidth]{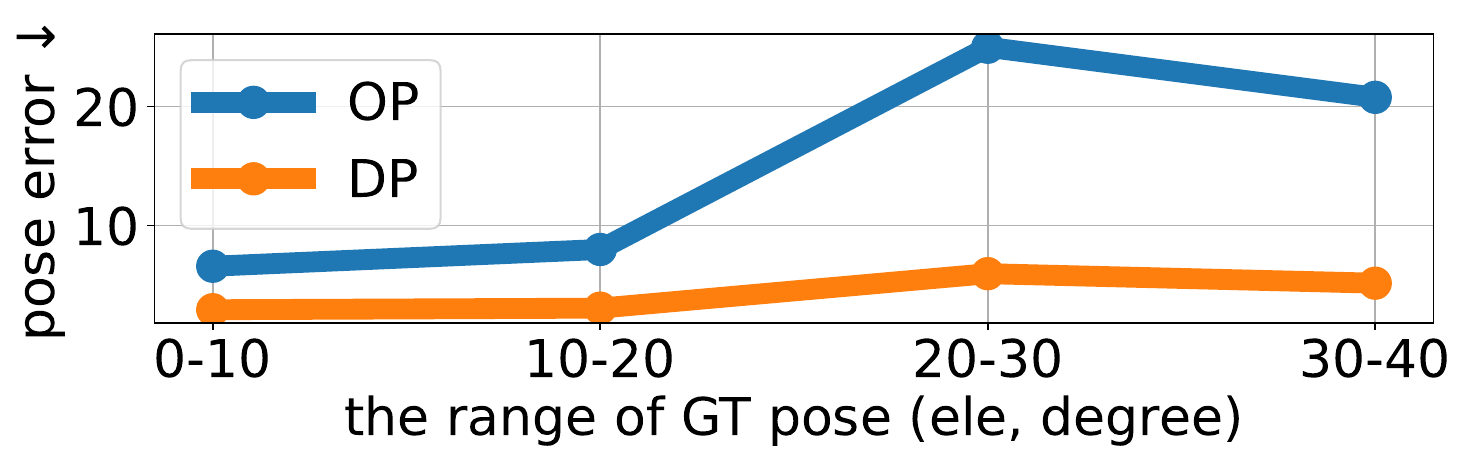}
    \\
    \includegraphics[width=.6\linewidth]{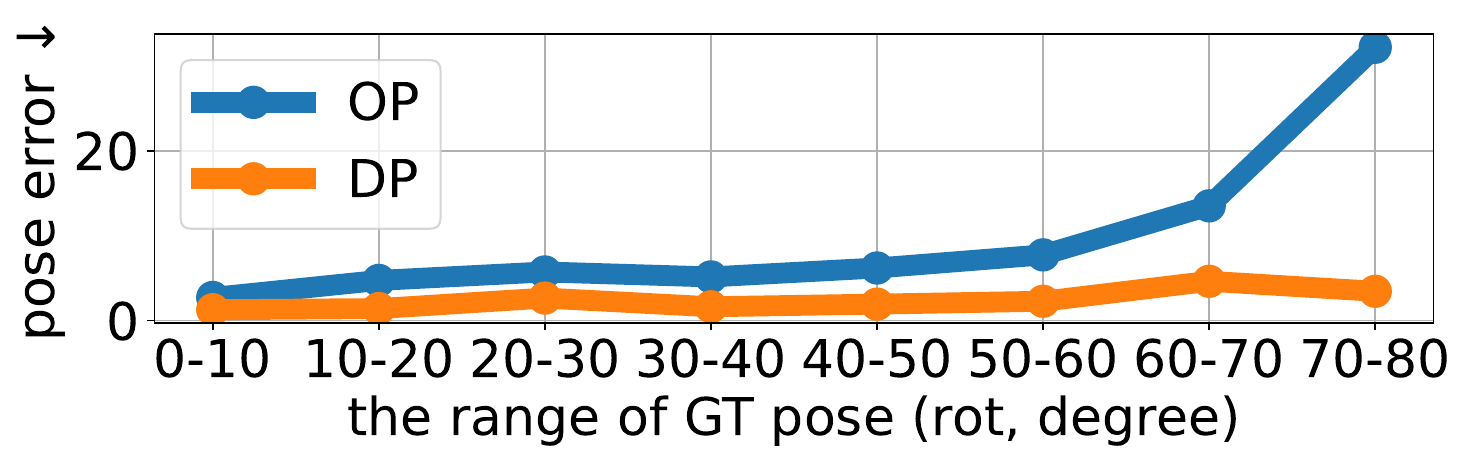}
  \end{tabular}
  \caption{
  Comparison of the pose reflection between ControlNet-DP and ControlNet-OP based on the estimated pose error between the reference images sampled from FFHQ~\cite{karras2019style} and the generated images by using conditions extracted from the reference images. Zero degrees of rotation and elevation indicate a frontal view. ControlNet-DP demonstrates lower error in both elevation and rotation compared to OP.
  Left: Pose error of ControlNet compared to SDEdit~\cite{meng2022sdedit} for reference. Each marker of the blue line indicates the pose error according to $t_0$, which represents the time step for adding noise to the input image in SDEdit. Right: Elevation(top) and rotation(bottom) error according to each range of ground truth pose.}
  \label{fig:cn_pose_error}
  \vspace{-5mm}
\end{figure}

To demonstrate the superiority of depth-based pose control, we compare its accuracy in pose control against the commonly used keypoints, generally obtained from OpenPose (OP).
For this, we meticulously assess the accuracy of pose reflection from the reference image to the generated image across two conditions. 
To be specific, we generate images using either OP or depth (DP) extracted from the given reference images and then compare the poses between the generated and provided images utilizing an off-the-shelf pose estimation model~\cite{deng2019accurate}. For this, we randomly sample 100 images from FFHQ~\cite{karras2019style} with a uniform pose distribution, and use them as reference images. From each condition extracted from the reference image, we generate 10 images to evaluate the pose reflection.
As depicted in the left graph of \cref{fig:cn_pose_error}, employing the DP as input of ControlNet for pose control better preserves the given pose compared to using the OP as input. Furthermore, as demonstrated in the right graphs of \cref{fig:cn_pose_error}, utilizing the DP as input consistently reflects the given poses across various pose ranges, while the pose error increases dramatically as the view moves away from the frontal view when using the OP as input of ControlNet. 

\section{Conclusion}
\label{sec:conclusion}

In this paper, we propose Skip-and-Play to generate images reflecting given poses across various objects. Specifically, we introduce depth-based pose control as opposed to the keypoints or camera parameters used in previous works for two reasons: 1) depth maps can be effortlessly obtained regardless of objects or poses, and 2) depth conditions inherently encode 3D spatial information, making them beneficial for controlling pose accurately in 3D space. However, the usage of the depth condition for pose control positions a challenge as it influences both the pose and shape of the generated images. To address this, we analyze the influence of the three components of the depth-conditional ControlNet on the shape and pose of generated images: 1) time steps using ControlNet, 2) ControlNet features obtained from negative prompts, and 3) ControlNet features passed to each decoder block. Based on empirical insights from the analysis, we design SnP by selectively skipping a part of three components.

Our experimental results demonstrate that SnP outperforms diffusion-based pose control models, qualitatively and quantitatively. While previous models are limited to generating images for specific objects or a restricted range of poses, SnP generates images across various objects and poses.

Our model is not free from limitations caused by leveraging the prior knowledge of ControlNet for pose-preserved image generation. Specifically, poses that are not adequately represented in ControlNet remain challenging for SnP to accurately express. This limitation arises from using ControlNet without additional training, but it can be mitigated as the performance of ControlNet improves.

%%%%%%%%% REFERENCES
{\small
\bibliographystyle{ieee_fullname}
\bibliography{egbib}
}

\end{document}